\documentclass[cameraready]{Interspeech}

\usepackage{multirow}
\usepackage{array}
\usepackage{tikz}
\usepackage{pgfplots}
\pgfplotsset{compat=1.18}
\definecolor{cbblack}{RGB}{0,0,0}
\definecolor{cbblue}{RGB}{0,114,178}
\definecolor{cborange}{RGB}{230,159,0}
\definecolor{cbgreen}{RGB}{0,158,115}
\definecolor{cbvermillion}{RGB}{213,94,0}
\definecolor{cbsky}{RGB}{86,180,233}
\definecolor{wongblue}{RGB}{0,114,178}
\definecolor{wongorange}{RGB}{230,159,0}
\definecolor{wonggreen}{RGB}{0,158,115}
\definecolor{wongvermillion}{RGB}{213,94,0}
\definecolor{wongsky}{RGB}{86,180,233}
\pgfplotsset{
  /pgfplots/xbar legend/.style={
    legend image code/.code={%
      \fill[##1] (0cm,-0.1cm) rectangle (0.3cm,0.2cm);%
    },
  },
}
\pgfplotsset{
  degradation/.style={
    width=52mm,
    height=4.8cm,
    xmin=-0.3, xmax=4.3,
    xtick={0,1,2,3,4},
    xticklabels={Clean,15,10,5,0},
    grid=major,
    grid style={gray!25},
    title style={font=\footnotesize},
    every axis plot/.append style={line width=0.7pt, mark size=1.5pt},
    clip=false,
  },
  probeleft/.style={
    width=52mm,
    height=4.8cm,
    xmin=-1, xmax=32,
    xtick={0,4,8,12,16,20,24,28,31},
    axis y line*=left,
    ymin=0, ymax=1.05,
    ytick={0,0.2,0.4,0.6,0.8,1.0},
    title style={font=\footnotesize},
    every axis plot/.append style={line width=0.7pt, mark size=1.5pt},
    clip=false,
  },
  proberight/.style={
    width=52mm,
    height=4.8cm,
    xmin=-1, xmax=32,
    xtick=\empty,
    axis y line*=right,
    ymin=0, ymax=0.6,
    ytick={0,0.1,0.2,0.3,0.4,0.5},
    yticklabel style={/pgf/number format/fixed},
    every axis plot/.append style={line width=0.7pt, mark size=1.5pt},
    clip=false,
  },
}
\usepackage{dblfloatfix}  

\title{The Cascade Equivalence Hypothesis:\\When Do Speech LLMs Behave Like ASR$\rightarrow$LLM Pipelines?}

\author[orcid=0000-0002-9365-0162]{Jayadev}{Billa$^{*}$}

\address{San Jose CA, USA.}
\email{jbilla2004@gmail.com}

\keywords{speech LLM, cascade equivalence, interpretability, logit lens, concept erasure}

\begin{document}
\makeatletter
\newcommand{\bstctlcite}[1]{\@bsphack
  \@for\@citeb:=#1\do{%
    \edef\@citeb{\expandafter\@firstofone\@citeb}%
    \if@filesw\immediate\write\@auxout{\string\citation{\@citeb}}\fi}%
  \@esphack}
\makeatother
\bstctlcite{IEEEexample:BSTcontrol}
\maketitle
\ifcameraready
  \renewcommand{\thefootnote}{\fnsymbol{footnote}}
  \setcounter{footnote}{1}
  \footnotetext{Unaffiliated researcher; previously at ISI@USC, Yahoo, Nuance, and BBN.}
  \renewcommand{\thefootnote}{\arabic{footnote}}
  \setcounter{footnote}{0}
\fi

\begin{abstract}
Speech LLMs are widely understood to be better than ASR$\rightarrow$LLM cascades since they have access to the audio directly, and not just the transcript. In this paper, we present an evaluation methodology and a mechanistic interpretation of the observed behavior of speech LLMs. First, we introduce matched-backbone testing which separates out the behavior of the speech LLM from the reasoning capabilities of the underlying LLM. Second, we provide a mechanistic analysis of speech LLMs using logit lens and LEACE and show the literal transcript emerging from the LLM's hidden states and that text representations are causally necessary. We also show that in most deployed use cases, current speech LLMs are expensive cascades, and under noise, they are worse ones, with clean-condition advantages reversing by up to 7.6\% at 0\,dB.

\end{abstract}

\section{Introduction}

End-to-end speech large language models (speech LLMs) such as Qwen2-Audio~\cite{chu2024qwen2audio}, Ultravox~\cite{ultravox2024}, Phi-4-Multimodal~\cite{phi4mm2025}, and Gemini~\cite{gemini2024} accept audio directly and produce text responses, bypassing the traditional pipeline of automatic speech recognition followed by a text LLM. The underlying promise being that the raw audio contains information (prosody, emotion, emphasis) that an ASR transcript cannot provide. This promise has led to substantial investment across diverse architectures, from discrete audio tokens~\cite{zhang2023speechgpt} to learned connectors~\cite{ultravox2024} to cross-attention~\cite{chu2024qwen2audio} to dual encoders~\cite{hu2024wavllm}, but it is not clear if the raw audio approach translates to genuinely different internal processing.

But does this promise hold? Or do they converge on implicit text representations, effectively becoming cascades with extra steps?

Broad benchmarks~\cite{huang2024dynamic,yang2024airbench} generally report aggregate performance only. Studies, such as LISTEN~\cite{chen2025listen} show that speech LLMs rely on the transcript even for emotion recognition, but did not investigate cascade comparison or analyze errors at an example level. In a related study, Cuervo et al.~\cite{cuervo2025salad} introduce the \emph{text understanding gap}, which describes the performance drop observed when a speech LLM processes spoken inputs relative to when the original text-based LLM processes the equivalent text; here, the LLM backbone is by definition the same, however, they compare audio input vs the original text input rather than an ASR transcript input which may have transcription errors. 

Separately, Cuervo and Marxer~\cite{cuervo2024scaling} show speech LLMs scale orders of magnitude less efficiently than text LLMs. This raises the question of whether direct speech processing is worth the investment when a cascade might suffice. Similar aggregate accuracy does not imply similar processing: two systems may achieve the same score through different per-example decisions. Distinguishing shared architecture from coincidentally similar performance requires per-example comparison, especially on errors. For example, if Qwen2-Audio (built on Qwen2-7B) is compared with a cascade of Whisper and Qwen2.5-7B, any difference between the two is a mixture of differences in audio processing architecture and differences in reasoning architecture. The model may “diverge” from the cascade simply because the backbone reasons differently, not because the audio is processed differently.

In this paper, we introduce \emph{matched-backbone behavioral testing} to address this issue. Whisper is paired with the \emph{same} LLM backbone as in each speech LLM (Llama-3.1-8B for Ultravox, Qwen2-7B for Qwen2-Audio, Phi-4-mini for Phi-4-Multimodal); this provides a controlled comparison that decouples architectural effects from backbone effects. We show that the backbone confound can inflate apparent architectural divergence by up to $+0.13\,\kappa$, closing much of the gap with the corresponding cascade equivalent.

We can formalize our inquiry as the \emph{Cascade Equivalence Hypothesis}: on tasks where the transcript has sufficient information to predict the task label, i.e., $I(A;\,Y \mid T) \approx 0$ with $A$ the audio, $T$ the transcript, and $Y$ the task label, a speech LLM and a cascade sharing the same LLM backbone should produce the same per-example responses, the same correct answers and the same errors, because both systems derive their predictions from the same textual information. We test this through Cohen's $\kappa$~\cite{cohen1960} for per-example agreement, conditional error overlap for shared failure modes, and McNemar's test~\cite{mcnemar1947} for systematic directional bias.

Our results establish cascade equivalence as a spectrum across architectures, with mechanistic evidence explaining why. The rest of the paper is organized as follows: methodology (\S2--3), behavioral findings (\S4), mechanistic evidence (\S5), and implications (\S6--7).

\noindent Our contributions are:
\begin{enumerate}
    \item \textbf{Matched-backbone behavioral testing}, a method for disentangling architecture vs. backbone artifacts in speech LLM benchmarking, revealing that the backbone confound inflates apparent architectural divergence by up to $+0.13\,\kappa$.
    \item \textbf{Empirical characterization of the cascade equivalence spectrum} across four speech LLMs, five cascades, and six tasks spanning text-sufficient and text-insufficient domains.
    \item \textbf{Mechanistic evidence across two architectures}: probing, logit lens, and causal erasure showing that speech LLMs build causally required text representations with architecture-dependent encoding that explains the behavioral equivalence spectrum.
    \item \textbf{Boundary conditions on equivalence}: cascade equivalence only holds under clean conditions; under noise, Whisper-based cascades substantially outperform all E2E models tested.
\end{enumerate}

\section{The Cascade Equivalence Hypothesis}

Consider a spoken input $A$, its transcript $T = \text{ASR}(A)$, and a task label $Y$.
We define the \emph{acoustic surplus} for task $Y$ as:
\begin{equation}
    \Delta I_Y = I(A;\,Y) - I(T;\,Y)
\end{equation}
When $\Delta I_Y \approx 0$, the transcript preserves nearly all task-relevant information; we call such tasks \emph{text-sufficient} (e.g., factual QA, topic classification, sentiment analysis).
Tasks for which the label depends on prosody that the transcription does not capture (e.g., emotion recognition, sarcasm detection) are \emph{text-insufficient}.
Text sufficiency is dependent on \emph{both} the model and task; a model may make errors of sufficient import that needed information is incorrect or incomplete. Billa~\cite{billa2021signal} demonstrated this empirically, showing that even low quality (i.e., high-WER) ASR transcripts in mismatched languages contain enough structure for downstream performance improvements. 

When $\Delta I_Y \approx 0$, the transcript includes all task-relevant information and any model that recovers $T$ from $A$, regardless of methodology, will necessarily produce the same downstream predictions. We define this as the \emph{Cascade Equivalence Hypothesis}: on text-sufficient tasks ($I(A;\,Y \mid T) \approx 0$), a speech LLM should behave identically to a cascade using the same LLM backbone. We highlight the fact that without the matched-backbone requirement, LLM reasoning capabilities cannot be disentangled from system performance.

This hypothesis lends itself to testable implications at two levels.
\emph{Behaviorally}, a speech LLM and its cascade equivalent should show high per-example agreement on text-sufficient tasks. This agreement should decrease on text-insufficient tasks if acoustic surplus is exploited. In cases where both systems are incorrect, they should predict the same wrong answer at rates above chance; this is the key signature of a shared reasoning pathway.
\emph{Mechanistically}, internal representations should progressively converge toward text, and erasing the text-predictive subspace should destroy task performance, confirming that text representations are causally necessary rather than epiphenomenal.

\section{Experimental Setup}

\subsection{Systems Under Test}

We evaluate four end-to-end speech LLMs and five ASR$\rightarrow$LLM cascades (Table~\ref{tab:systems}).
All cascades use Whisper-large-v3~\cite{radford2023whisper} for ASR.
Three ``matched-backbone'' cascades pair Whisper with the \emph{same} LLM backbone used inside each open-weight speech LLM: Qwen2-7B~\cite{yang2024qwen25} for Qwen2-Audio, Llama-3.1-8B~\cite{dubey2024llama3} for Ultravox, and Phi-4-mini for Phi-4-Multimodal.
Two additional cascades, ``strong'' (Whisper-large\,$\to$\,Qwen2.5-7B) and ``weak'' (Whisper-small\,$\to$\,Qwen2.5-7B), provide upper and lower reference points.

\begin{table}[t]
\centering
\caption{Systems evaluated.
$^\dagger$Matched-backbone cascades share the LLM with the corresponding speech LLM.}
\label{tab:systems}
\footnotesize
\setlength{\tabcolsep}{2.5pt}
\begin{tabular}{llr}
\toprule
\textbf{System} & \textbf{Type} & \textbf{Params} \\
\midrule
Whisper-large $\to$ Qwen2.5-7B & Cascade (strong) & 1.5B+7B \\
Whisper-small $\to$ Qwen2.5-7B & Cascade (weak)   & 244M+7B \\
Whisper-large $\to$ Qwen2-7B$^\dagger$   & Matched cascade  & 1.5B+7B \\
Whisper-large $\to$ Llama-3.1-8B$^\dagger$   & Matched cascade  & 1.5B+8B \\
Whisper-large $\to$ Phi-4-mini$^\dagger$  & Matched cascade  & 1.5B+3.8B \\
\midrule
Qwen2-Audio-7B     & End-to-end       & 8.2B \\
Ultravox v0.6      & End-to-end       & 8.4B \\
Phi-4-Multimodal   & End-to-end       & 5.6B \\
Gemini 2.0 Flash   & End-to-end (API) & --- \\
\bottomrule
\end{tabular}
\end{table}

\subsection{Evaluation Tasks}

We evaluate on six tasks spanning the text-sufficient to text-insufficient spectrum. \textbf{Text-sufficient tasks} are: \emph{TriviaQA}~\cite{joshi2017triviaqa}, open-ended factual QA (accuracy table only; excluded from behavioral analysis due to open-ended answers);
\emph{AG News}~\cite{zhang2015character}, 4-way topic classification;
\emph{SST-2}~\cite{socher2013sst}, binary sentiment analysis;
\emph{CommonsenseQA}~\cite{talmor2019commonsenseqa}, 5-way commonsense reasoning.
For text-sufficient tasks, we use TTS-synthesized voices with six Microsoft Edge-TTS\footnote{\url{https://github.com/rany2/edge-tts}}. \textbf{Text-insufficient tasks} are: 
\emph{MELD}~\cite{poria2019meld}, 7-way emotion recognition from \textit{Friends} dialogue (2{,}610 utterances);
\emph{MUStARD}~\cite{castro2019mustard}, binary sarcasm detection from TV shows (690 clips). All text-insufficient tasks use natural voice content.
The total test is 6{,}300 evaluation samples, 27{,}300 audio files. Noise perturbation conditions (multi-talker babble at 15, 10, 5, and 0\,dB SNR using MUSAN~\cite{snyder2015musan}) are applied to two TTS voices across the text-sufficient tasks.

\subsection{Behavioral Metrics}

For each (speech LLM, cascade) pair on each classification task, we compute: \emph{Cohen's $\kappa$}, chance-corrected per-example agreement;
\emph{conditional error overlap}, $P(\text{same wrong answer} \mid \text{both wrong})$, testing shared failure mechanisms;
and \emph{McNemar's test}, whether one system is systematically better (Benjamini-Hochberg FDR-corrected across all pair$\times$task comparisons).

\subsection{Probing and Intervention}

For the two open-weight speech LLMs (Qwen2-Audio, Ultravox) and Whisper (as a control), we extract hidden states at nine layers ($\ell \in \{0, 4, 8, 12, 16, 20, 24, 28, 31\}$) on 1{,}000 utterances and train:
\textit{acoustic probes}, linear regression for energy and pitch, scored by coefficient of determination ($R^2$); 
\textit{text probes}: a CTC probe, a per-frame linear classifier trained with CTC loss to decode character sequences from hidden states, scored by text decodability ($1{-}\text{CER}$, where higher values indicate more decodable text); and a bag-of-characters (BoC) probe, linear regression on character frequency vectors ($R^2$).
This linear probing methodology~\cite{alain2017probing} is standard for high-dimensional representations ($d{\approx}4096$; 80/20 train/test split).

\noindent\textit{Logit lens}~\cite{nostalgebraist2020logitlens} projects hidden states at audio token positions (for Ultravox) or all tokens (for Qwen2-Audio) through the LLM's own unembedding matrix at each layer.
Following standard practice for pre-norm architectures~\cite{belrose2023tuned}, we apply the model's final RMSNorm to intermediate hidden states before projecting through the unembedding matrix, so that the scale and distribution of hidden states matches what the language modeling head expects.
If top-predicted tokens match transcript words, the model is literally transcribing audio to text internally.
AudioLens~\cite{yang2025audiolens} recently applied similar projection to audio-language models for attribute recognition; we apply the technique to study text emergence in speech LLMs.

\noindent\textit{LEACE} (LEAst-squares Concept Erasure)~\cite{belrose2024leace} provides causal evidence by surgically removing specific information from the model's representations during inference.
We use the \texttt{concept-erasure} library\footnote{\url{https://github.com/EleutherAI/concept-erasure}} to compute optimal linear erasers from hidden states and concept labels at each layer.
If the model relies on text information for its predictions, erasing the text-predictive directions should collapse performance; if it does not, erasure should have no effect.
We apply erasure at all probed layers simultaneously on every forward pass, including during autoregressive generation, to ensure that the model cannot recover the erased information from unmodified layers~\cite{elazar2021amnesic,dobrzeniecka2025amnesic}.
We disable bias centering (\texttt{use\_bias=False}) because the training mean, computed from audio-token hidden states, is out-of-distribution for text tokens generated during decoding, causing model collapse when applied.
Since text specificity is dependent on both task and model, we test across multiple erasure conditions: \emph{BoC} fits LEACE to character frequency vectors (48-dimensional concept); \emph{proxy erasure} fits LEACE to first-word classification labels (159-dimensional concept); \emph{CTC erasure} fits LEACE to per-frame character predictions from the trained CTC probe (49-class concept, capturing sequential character order).

Finally, we add two control conditions: \emph{random erasure} removes a random 159-dimensional orthogonal subspace as a dimensionality-matched control; \emph{acoustic erasure} fits LEACE on pitch and energy values (2-dimensional concept).

\section{Behavioral Results}

\begin{table}[t]
\centering
\caption{System accuracy (\%) on clean audio, classification tasks.
TriviaQA accuracy (open-ended): Cascade-S 38.8, Cascade-W 36.2, Ultravox 37.5, Qwen2-Audio 8.2, Phi-4-MM 14.9, Gemini 48.4.
Qwen2-Audio's low TriviaQA score reflects its older Qwen2-7B backbone's weak factual recall, not an audio processing failure; on classification tasks it is competitive.}
\label{tab:accuracy}
\footnotesize
\setlength{\tabcolsep}{3pt}
\begin{tabular}{l rrrrr}
\toprule
\textbf{System} & \textbf{AG} & \textbf{SST} & \textbf{CSQA} & \textbf{MELD} & \textbf{MUS} \\
\midrule
Cascade-S & 85.0 & 88.5 & 79.3 & 51.8 & 60.0 \\
Cascade-W & 85.0 & 87.8 & 78.7 & 51.3 & 59.6 \\
\addlinespace
Ultravox       & 83.3 & 85.3 & 70.5 & 39.8 & 57.7 \\
Qwen2-Audio    & 80.5 & 82.2 & 62.8 & 43.6 & 52.8 \\
Phi-4-MM       & 81.5 & 82.0 & 61.1 & 30.2 & 61.0 \\
Gemini         & 83.6 & 89.7 & 80.2 & 51.9 & 68.1 \\
\bottomrule
\end{tabular}
\end{table}

\subsection{Agreement on Text-Sufficient Tasks}
\label{sec:agreement}

Table~\ref{tab:accuracy} presents accuracy and Figure~\ref{fig:kappa_heatmap} shows matched-backbone $\kappa$ for all systems. On text-sufficient tasks, cascades match or exceed every open-weight E2E model in accuracy, despite their simpler architecture. Only Gemini, a closed-weight system, matches the cascade on AG News and exceeds it on SST-2 and CSQA. This accuracy parity already suggests that the E2E models may not be extracting additional value from the raw audio on these tasks. On text-insufficient tasks (MELD, MUStARD), all open-weight E2E models underperform the cascades, suggesting that direct audio access is not yet translating into better paralinguistic processing. But do these similar scores come from the same per-example decisions, or just happen to average out the same way?

\textit{Ultravox is near-cascade-equivalent.}
Against its matched Llama-3.1-8B cascade, Ultravox achieves $\kappa = 0.93$ [0.92, 0.94] on AG News (95\% bootstrap CIs throughout), $0.78$ [0.76, 0.79] on CSQA, and $0.75$ [0.74, 0.77] on SST-2. On CSQA, McNemar's test is non-significant after FDR correction ($p = 0.29$): the systems make errors at statistically indistinguishable rates. The CSQA $\kappa$ jumps from $0.65$ (against the mismatched Cascade-S) to $0.78$ with the matched backbone, a $+0.13$ increase that closes much of the gap toward the cascade ceiling ($0.96$).

\textit{Qwen2-Audio genuinely diverges.}
Against its Qwen2-7B backbone cascade, $\kappa$ on text-sufficient tasks remains moderate ($0.54$--$0.85$) and shows minimal improvement over the mismatched comparison. On SST-2, $\kappa$ actually \emph{decreases} from $0.62$ to $0.54$. This suggests that Qwen2-Audio processes sentiment in a way that is architecturally distinct from a cascade.

The cascade-S vs.\ cascade-W ceiling validates our text-sufficient/insufficient taxonomy: $\kappa = 0.93$--$0.98$ despite a substantial ASR quality gap, confirming that even degraded transcripts carry nearly all task-relevant information.
\textit{Phi-4-MM and Gemini lie in between these extremes}, showing that cascade equivalence is not a binary, but rather a continuous property (Figure~\ref{fig:kappa_heatmap}). Phi-4-MM's Mixture-of-LoRAs architecture achieves $\kappa{=}0.85$ on AG News (matching Qwen2-Audio) but only $0.23$ on MELD, the lowest of any matched pair. Combined with its low MELD accuracy (30.2\%, Table~\ref{tab:accuracy}), this indicates that Phi-4-MM does not adhere to the cascade text pathway, and does not succeed in extracting useful paralinguistic information either; its modality router appears to simply discard emotion-relevant acoustic information.

\begin{figure*}[!tb]
    \centering
    \input{fig_kappa_heatmap.tex}
    \caption{Cohen's $\kappa$ between each E2E model and its matched-backbone cascade ($^\dagger$) or Cascade-S.
    Rows are ordered by decreasing mean $\kappa$.
    95\% bootstrap CIs (1{,}000 resamples) are ${\pm}0.01$--$0.02$ on text-sufficient tasks and ${\pm}0.03$--$0.07$ on MELD/MUStARD.
    Ultravox shows consistently high agreement, while Qwen2-Audio and Phi-4-MM show lower and more variable agreement.}
    \label{fig:kappa_heatmap}
\end{figure*}

\subsection{Shared Error Signatures}

If cascade equivalence holds, the two systems should not only agree on correct answers but also fail on the \emph{same} examples and produce the \emph{same} wrong answers. We measure this with conditional error overlap: $P(\text{same wrong answer} \mid \text{both wrong})$. Since SST-2 is binary (only one possible wrong answer), we report this metric only for multi-class tasks.

\begin{figure}[!htb]
    \centering
\begin{tikzpicture}
\begin{axis}[
    xbar=0.4pt,
    bar width=4.5pt,
    width=\columnwidth,
    height=5.5cm,
    font=\footnotesize,
    xlabel={Conditional Error Overlap},
    xlabel style={font=\footnotesize},
    xmin=0, xmax=1.15,
    xtick={0, 0.2, 0.4, 0.6, 0.8, 1.0},
    ytick=data,
    symbolic y coords={MELD, CSQA, AG News},
    y tick label style={font=\footnotesize},
    x tick label style={font=\scriptsize},
    enlarge y limits=0.15,
    legend style={
        at={(0.5,1.02)},
        anchor=south,
        font=\tiny,
        legend columns=3,
        column sep=3pt,
        draw=none,
        fill=none,
        /tikz/every even column/.append style={column sep=3pt},
    },
    clip=false,
    every axis plot/.append style={fill opacity=0.90, draw=none},
    nodes near coords,
    every node near coord/.append style={font=\tiny, anchor=west},
]

\addplot[fill=wongsky]
    coordinates {(0.94,AG News) (0.68,CSQA) (0.44,MELD)};

\addplot[fill=wongvermillion]
    coordinates {(0.96,AG News) (0.85,CSQA) (0.68,MELD)};

\addplot[fill=wonggreen]
    coordinates {(0.95,AG News) (0.70,CSQA) (0.62,MELD)};

\addplot[fill=wongorange]
    coordinates {(0.87,AG News) (0.61,CSQA) (0.48,MELD)};

\addplot[fill=wongblue]
    coordinates {(0.86,AG News) (0.59,CSQA) (0.52,MELD)};

\legend{Phi-4-MM (matched), Ultravox (matched), Ultravox vs Cas-S, Qwen2-Audio (matched), Qwen2-Audio vs Cas-S}

\draw[black, dashed, line width=0.5pt]
    ([yshift=-12pt]axis cs:0.333,AG News) -- ([yshift=12pt]axis cs:0.333,AG News);
\draw[black, dashed, line width=0.5pt]
    ([yshift=-12pt]axis cs:0.25,CSQA) -- ([yshift=12pt]axis cs:0.25,CSQA);
\draw[black, dashed, line width=0.5pt]
    ([yshift=-12pt]axis cs:0.167,MELD) -- ([yshift=12pt]axis cs:0.167,MELD);

\end{axis}
\end{tikzpicture}
    \caption{Conditional error overlap: $P(\hat{y}_{e2e}{=}\hat{y}_{cas} \mid \text{both wrong})$ for each speech LLM vs.\ its cascade counterpart on multi-class tasks.
    Values near 1.0 mean both systems produce the \emph{same} wrong answer when they err.
    Dashed lines mark chance baselines ($1/(|C|{-}1)$).
    Matched-backbone pairs consistently achieve higher overlap than mismatched pairs (vs Cas-S).}
    \label{fig:error_overlap}
\end{figure}
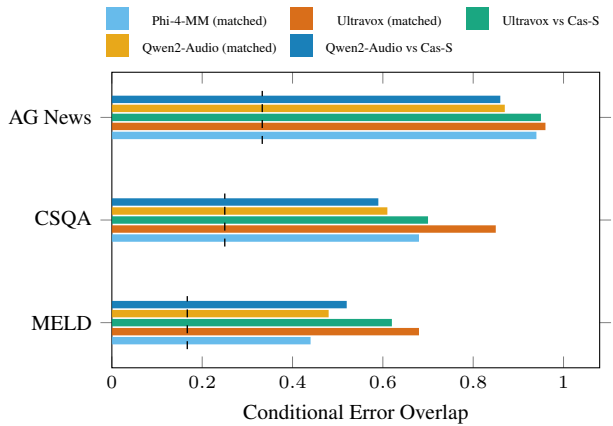

\textit{Ultravox and its matched cascade share nearly identical failure modes.}
On AG~News, conditional error overlap between Ultravox and its matched Llama-3.1-8B cascade is 0.96---when both systems err, they almost always pick the same wrong label (Figure~\ref{fig:error_overlap}). Of these joint errors, 81\% occur despite a correct transcription: both systems decoded the audio perfectly but failed at the same classification boundary. The audio pathway played no role; these are LLM reasoning failures.
On CSQA the overlap is 0.85; on MELD it drops to 0.68, in line with the greater behavioral divergence we expect on text-insufficient tasks.
These rates are $2.9$--$4.1{\times}$ the chance baselines ($1/(|C|{-}1)$).

\textit{Other matched pairs show the same pattern.}
Across all models, matched-backbone pairs consistently achieve higher overlap than their mismatched counterparts, which tells us the shared failures come from the LLM backbone, not the audio encoder.
Phi-4-MM's matched pair reaches 0.94 on AG~News but only 0.44 on MELD, again tracking its weaker agreement on text-insufficient tasks.

Table~\ref{tab:error_examples} shows representative shared errors between Ultravox and its matched cascade.
We see two failure modes: \emph{LLM reasoning errors}, where both systems transcribe correctly but land on the same wrong class (e.g., tech-company stories labeled Business instead of Sci/Tech), and \emph{ASR-induced errors}, where identical transcription mistakes produce identical downstream failures (e.g., ``mountie'' misrecognized as ``Amante,'' destroying the Canada clue).
On SST-2, 66\% of shared errors involve a degraded transcript, which is understandable since a single dropped negation or intensifier can flip the sentiment.

\begin{table}[t]
\centering
\caption{Shared error examples: both Ultravox and its matched cascade produce the identical wrong answer.
Same-answer rates when both err: AG News 96\%, CSQA 85\%.}
\label{tab:error_examples}
\footnotesize
\setlength{\tabcolsep}{3pt}
\begin{tabular}{p{4.5cm}ccc}
\toprule
\textbf{Input (abridged)} & \textbf{Gold} & \textbf{Both} & \textbf{Type} \\
\midrule
\multicolumn{4}{l}{\textit{AG News}} \\
``Cisco \& Microsoft Partner for CRM'' & Sci/Tech & Bus. & LLM \\
``iTunes selling Band Aid song'' & Sci/Tech & Bus. & LLM \\
\addlinespace
\multicolumn{4}{l}{\textit{CSQA}} \\
``A mountie at a subway'' $\to$ ``Amante...'' & Toronto & NYC & ASR \\
``What do people do playing guitar?'' & Singing & Music & LLM \\
\addlinespace
\multicolumn{4}{l}{\textit{SST-2}} \\
``chilling tale of great crimes'' & Pos. & Neg. & Ambig. \\
\bottomrule
\end{tabular}
\end{table}

\subsection{Text-Insufficient Tasks}

On MELD and MUStARD, $\kappa$ drops substantially for all pairs (Figure~\ref{fig:kappa_heatmap}), as the cascade equivalence hypothesis predicts for text-insufficient tasks.
However, using the matched-backbone comparison, we find that the backbone confound contributes significantly to the gap.
Ultravox's MELD $\kappa$ improves from 0.40 (vs.\ Cascade-S) to 0.52 (vs.\ its matched cascade); MUStARD from 0.25 to 0.55.
This demonstrates that even for text-insufficient tasks LLM backbones are a confound, and highlights the need for matched-backbone comparison. 
The matched $\kappa$ values (${\sim}0.5$) are well below the text-sufficient ceiling ($0.93$--$0.98$), consistent with greater divergence on tasks where acoustic information should matter.

Qwen2-Audio shows minimal change from backbone matching: MELD $\kappa$ moves from $0.37$ to $0.30$, and MUStARD from $0.05$ to $0.05$.
This is not unexpected given that the mismatched (Qwen2.5-7B) and matched (Qwen2-7B) backbones share the same Qwen lineage.
MUStARD $\kappa \approx 0.05$ (near zero) indicates either the model is using paralinguistic features or that it is using a qualitatively different processing strategy.
Since Qwen2-Audio's MUStARD accuracy (52.8\%) is barely above chance, a different but not better strategy is more likely.

\subsection{Noise Robustness}

Under multi-talker babble noise (MUSAN, 0--15\,dB SNR), our Whisper-based cascades are more robust than all tested E2E models (Figure~\ref{fig:degradation}). For example, at 0\,dB, Cascade-S loses 0.5--4.2\% accuracy across text-sufficient tasks, while E2E models show larger drops on SST-2 and CSQA (3.9--12.7\%). This is not surprising given that Whisper, trained on 680{,}000 hours of diverse audio, serves as a noise-robust front-end that absorbs acoustic degradation before it reaches the LLM.

Gemini degrades fastest despite the best clean accuracy: SST-2 drops $10.2$\% at 0\,dB (90.4\%\,$\to$\,80.2\%) vs.\ Cascade-S's 2.6\% drop.
This means that \emph{selecting a system based on clean-condition accuracy can reverse under deployment noise}: a team choosing Gemini over Cascade-S for SST-2 would see a 7.6 percentage-point reversal at 0\,dB (from $+$2.0\% clean advantage to $-$5.6\% disadvantage), invisible in standard benchmarks.

\begin{figure*}[!tb]
    \centering
\begin{tikzpicture}[font=\footnotesize]
\begin{axis}[
  degradation,
  at={(0,0)},
  ylabel={Accuracy (\%)},
  title={AG News},
  ymin=76, ymax=88,
  ytick={76,78,80,82,84,86,88},
  legend style={
    at={(0.5,1.02)},
    anchor=south,
    font=\scriptsize,
    legend columns=5,
    column sep=3pt,
    draw=gray!60, fill=white,
    rounded corners=1pt,
    inner sep=2pt,
  },
  legend to name=degradlegend,
]
\addplot[cbblack, mark=o] coordinates {(0,85.15)(1,84.75)(2,85.15)(3,84.75)(4,84.65)};
\addlegendentry{Cascade-S}
\addplot[cbblack, dashed, mark=x] coordinates {(0,85.0)(1,84.7)(2,84.9)(3,85.25)(4,83.75)};
\addlegendentry{Cascade-W}
\addplot[cbblue, mark=triangle] coordinates {(0,80.6)(1,79.85)(2,79.85)(3,79.6)(4,78.65)};
\addlegendentry{Qwen2-Audio}
\addplot[cborange, mark=diamond] coordinates {(0,83.1)(1,83.2)(2,83.0)(3,83.0)(4,82.8)};
\addlegendentry{Ultravox}
\addplot[cbgreen, mark=star] coordinates {(0,83.75)(1,83.2)(2,83.25)(3,83.2)(4,82.55)};
\addlegendentry{Gemini}
\end{axis}

\begin{axis}[
  degradation,
  at={(55mm,0)},
  title={SST-2},
  ymin=76, ymax=92,
  ytick={76,78,80,82,84,86,88,90,92},
]
\addplot[cbblack, mark=o] coordinates {(0,88.4)(1,88.9)(2,88.3)(3,87.8)(4,85.85)};
\addplot[cbblack, dashed, mark=x] coordinates {(0,87.2)(1,87.5)(2,87.1)(3,86.15)(4,82.6)};
\addplot[cbblue, mark=triangle] coordinates {(0,82.45)(1,81.95)(2,81.1)(3,81.1)(4,78.55)};
\addplot[cborange, mark=diamond] coordinates {(0,85.5)(1,84.5)(2,84.6)(3,83.75)(4,79.65)};
\addplot[cbgreen, mark=star] coordinates {(0,90.4)(1,88.15)(2,87.2)(3,85.5)(4,80.25)};
\end{axis}

\begin{axis}[
  degradation,
  at={(110mm,0)},
  title={CSQA},
  ymin=52, ymax=84,
  ytick={55,60,65,70,75,80},
]
\addplot[cbblack, mark=o] coordinates {(0,79.35)(1,78.85)(2,79.15)(3,78.4)(4,75.2)};
\addplot[cbblack, dashed, mark=x] coordinates {(0,78.5)(1,78.8)(2,78.25)(3,76.7)(4,69.3)};
\addplot[cbblue, mark=triangle] coordinates {(0,63.95)(1,60.9)(2,60.25)(3,60.1)(4,55.1)};
\addplot[cborange, mark=diamond] coordinates {(0,70.75)(1,70.0)(2,69.6)(3,67.8)(4,62.45)};
\addplot[cbgreen, mark=star] coordinates {(0,81.5)(1,78.8)(2,78.05)(3,76.7)(4,68.85)};
\end{axis}

\path (current bounding box.north west) -- (current bounding box.north east)
  node[midway, anchor=south, yshift=2pt] {\pgfplotslegendfromname{degradlegend}};
\end{tikzpicture}
    \caption{Accuracy vs.\ SNR for text-sufficient tasks.
    Cascade-S (solid line, circles) degrades gracefully across all tasks.
    Gemini (solid line, triangles) shows the steepest decline on SST-2 and CSQA despite superior clean performance, suggesting higher noise sensitivity in its internal speech processing.}
    \label{fig:degradation}
\end{figure*}
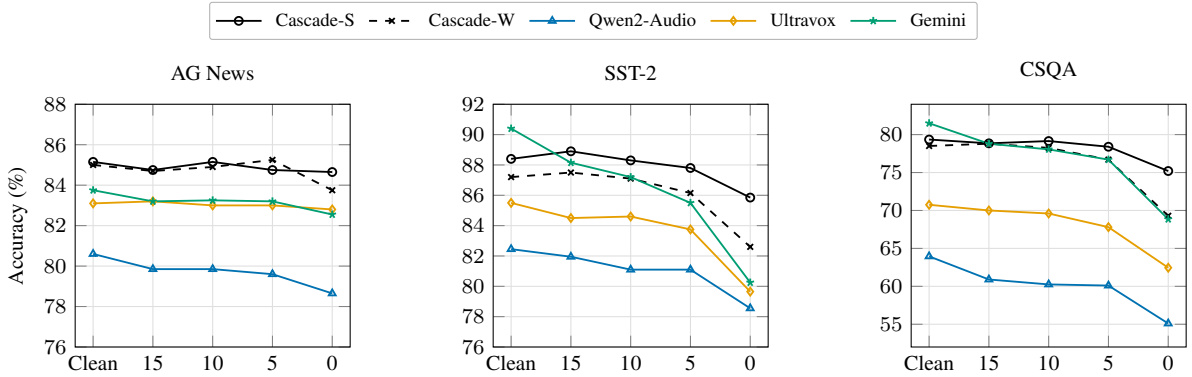

\begin{figure*}[!tb]
    \centering
\begin{tikzpicture}[font=\footnotesize]
\begin{axis}[
  probeleft,
  at={(0,0)},
  ylabel={Acoustic $R^2$},
  ylabel style={color=cbblue},
  y tick label style={color=cbblue},
  title={Qwen2-Audio},
  legend style={
    at={(0.5,1.02)},
    anchor=south,
    font=\scriptsize,
    legend columns=2,
    column sep=3pt,
    draw=gray!60, fill=white,
    rounded corners=1pt,
    inner sep=2pt,
  },
  legend to name=probelegend,
]
\addplot[cbblue, mark=o] coordinates {
  (0,0.9205)(4,0.9204)(8,0.9195)(12,0.9096)(16,0.8969)(20,0.8793)(24,0.8659)(28,0.8593)(31,0.8563)
};
\addlegendentry{Energy $R^2$}
\addplot[cbblue, dashed, mark=x, opacity=0.7] coordinates {
  (0,0.6269)(4,0.6247)(8,0.6063)(12,0.5985)(16,0.5735)(20,0.5381)(24,0.491)(28,0.4769)(31,0.4835)
};
\addlegendentry{Pitch $R^2$}
\end{axis}
\begin{axis}[
  proberight,
  at={(0,0)},
  y tick label style={color=cborange},
  legend style={
    at={(0.5,1.02)},
    anchor=south,
    font=\scriptsize,
    legend columns=2,
    column sep=3pt,
    draw=gray!60, fill=white,
    rounded corners=1pt,
    inner sep=2pt,
  },
  legend to name=proberightlegend,
]
\addplot[cborange, mark=triangle] coordinates {
  (0,0.502)(4,0.481)(8,0.489)(12,0.495)(16,0.510)(20,0.485)(24,0.433)(28,0.393)(31,0.291)
};
\addlegendentry{CTC decod.}
\addplot[cborange, dotted, mark=square*, mark size=1.5pt] coordinates {
  (0,0.188)(4,0.1988)(8,0.204)(12,0.1899)(16,0.1857)(20,0.193)(24,0.1768)(28,0.1637)(31,0.1492)
};
\addlegendentry{BoC $R^2$}
\end{axis}

\begin{axis}[
  probeleft,
  at={(55mm,0)},
  y tick label style={color=cbblue},
  title={Ultravox},
]
\addplot[cbblue, mark=o] coordinates {
  (0,0.8599)(4,0.858)(8,0.8555)(12,0.8572)(16,0.85)(20,0.8427)(24,0.832)(28,0.8342)(31,0.8347)
};
\addplot[cbblue, dashed, mark=x, opacity=0.7] coordinates {
  (0,0.368)(4,0.3603)(8,0.3489)(12,0.3601)(16,0.3602)(20,0.3627)(24,0.3436)(28,0.3407)(31,0.3121)
};
\end{axis}
\begin{axis}[
  proberight,
  at={(55mm,0)},
  y tick label style={color=cborange},
]
\addplot[cborange, mark=triangle] coordinates {
  (0,0.031)(4,0.037)(8,0.0)(12,0.031)(16,0.067)(20,0.079)(24,0.107)(28,0.175)(31,0.202)
};
\addplot[cborange, dotted, mark=square*, mark size=1.5pt] coordinates {
  (0,0.0749)(4,0.1234)(8,0.1203)(12,0.1317)(16,0.1564)(20,0.1557)(24,0.1505)(28,0.1476)(31,0.1181)
};
\end{axis}

\begin{axis}[
  probeleft,
  at={(105mm,0)},
  y tick label style={color=cbblue},
  title={Whisper (control)},
]
\addplot[cbblue, mark=o] coordinates {
  (0,0.9858)(4,0.9835)(8,0.983)(12,0.9826)(16,0.9764)(20,0.964)(24,0.9715)(28,0.9682)(31,0.9627)
};
\addplot[cbblue, dashed, mark=x, opacity=0.7] coordinates {
  (0,0.7126)(4,0.7446)(8,0.7803)(12,0.7886)(16,0.7566)(20,0.7528)(24,0.731)(28,0.7045)(31,0.6673)
};
\end{axis}
\begin{axis}[
  proberight,
  at={(105mm,0)},
  ylabel={Text Decod.},
  ylabel style={color=cborange},
  y tick label style={color=cborange},
]
\addplot[cborange, mark=triangle] coordinates {
  (0,0.2577)(4,0.2115)(8,0.1882)(12,0.2159)(16,0.1593)(20,0.2)(24,0.193)(28,0.2425)(31,0.2566)
};
\addplot[cborange, dotted, mark=square*, mark size=1.5pt] coordinates {
  (0,0.0585)(4,0.0763)(8,0.0877)(12,0.0994)(16,0.1059)(20,0.1297)(24,0.1165)(28,0.133)(31,0.1364)
};
\end{axis}

\path (current bounding box.north west) -- (current bounding box.north east)
  node[pos=0.35, anchor=south, yshift=2pt] {\pgfplotslegendfromname{probelegend}}
  node[pos=0.70, anchor=south, yshift=2pt] {\pgfplotslegendfromname{proberightlegend}};
\end{tikzpicture}
    \caption{Layer-wise probing.
    \textbf{Left axis}: acoustic probe $R^2$ for energy (solid, circles) and pitch (dashed, squares).
    \textbf{Right axis}: CTC text decodability (solid, triangles) and bag-of-characters $R^2$ (dashed, diamonds).
    Qwen2-Audio shows the strongest acoustic compression; Ultravox retains acoustics while building text representations.}
    \label{fig:probes}
\end{figure*}
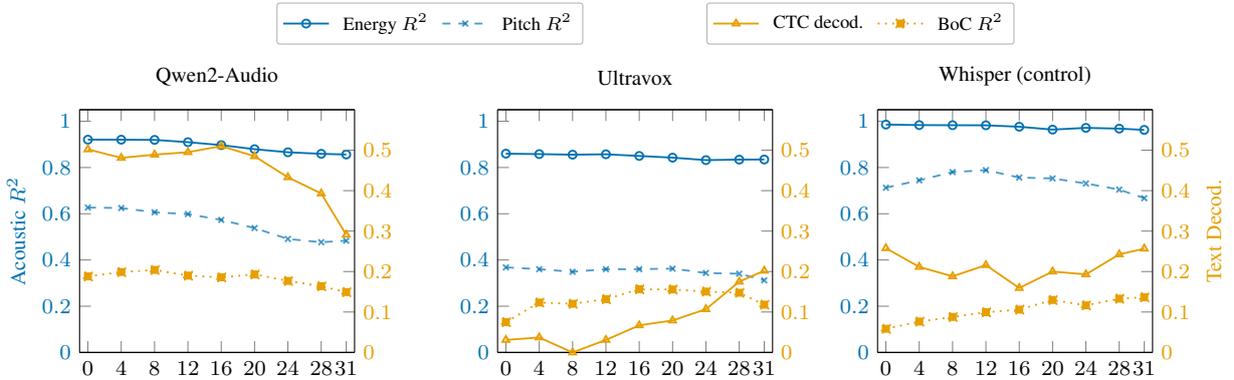

\section{Mechanistic Evidence}

Given these behavioral results, we dig deeper and ask what is encoded (probing), how it emerges (logit lens), whether the generated text is task-informative (functional text test), and whether the text is causally required (LEACE).

\subsection{Layer-wise Probing} \label{sec:layer-wise-probe}

Figure~\ref{fig:probes} and Table~\ref{tab:probes} show acoustic and text probe results for Qwen2-Audio, Ultravox, and Whisper as a control.
Recall that Whisper, as an ASR model, is designed specifically to encode input audio into text, so it acts as a useful baseline to understand how acoustic and text representations should be distributed across layers in such an encoder. Indeed, the decodability of the CTC text representation in the layers of Whisper follows a non-monotonic curve, falling to 0.16 at layer 16 and then rising again to 0.26 at layer 31, suggesting an encoding bottleneck in which mid-layers focus on acoustic representation and later layers focus on text encoding.

\textit{Acoustic information is preserved but transformed.}
This is evidenced by the fact that Energy $R^2$ does not drop dramatically for any of the three models: 0.92\,$\to$\,0.86 for Qwen2-Audio, 0.86\,$\to$\,0.83 for Ultravox, and 0.99\,$\to$\,0.96 for Whisper. In contrast, pitch $R^2$ falls more drastically (Qwen2-Audio: 0.63\,$\to$\,0.48;  Ultravox: 0.37\,$\to$\,0.31), indicating that more detailed acoustic characteristics are being increasingly lost as the model generates text representations.

\textit{Text decodability reveals divergent architectures.}
Ultravox shows progressive text emergence (CTC: 0.03\,$\to$\,0.20), while Qwen2-Audio maintains stable text decodability (0.48--0.51) through layers 0--20 before a sharp decline in the final layers (0.43 at L24, 0.29 at L31). Qwen2-Audio's cross-attention encoder delivers text-decodable representations to the LLM from the outset (TD${\approx}$0.50 at L0), and only declines in later layers as the model shifts toward task reasoning. Ultravox's connector, in contrast, passes representations with almost no text structure (TD$=$0.03 at L0); the LLM itself must progressively build text decodability through its layers. This difference mirrors the division of labor in a cascade: Qwen2-Audio's encoder acts like the ASR stage, while Ultravox distributes the transcription work across the full model.

The \emph{acoustic compression gradient maps to behavioral agreement}: Qwen2-Audio (steepest pitch decline, 0.63\,$\to$\,0.48) shows the lowest $\kappa$ with its cascade; Ultravox (milder decline, 0.37\,$\to$\,0.31) shows higher $\kappa$.

\begin{table}[t]
\centering
\caption{Layer-wise probe results at selected layers.
Energy/Pitch: ridge regression $R^2$; CTC: text decodability (fraction of characters correctly recovered); BoC: bag-of-characters $R^2$ (ridge).}
\label{tab:probes}
\footnotesize
\setlength{\tabcolsep}{3pt}
\begin{tabular}{clcccc}
\toprule
& \textbf{Layer} & \textbf{Energy} & \textbf{Pitch} & \textbf{CTC} & \textbf{BoC} \\
\midrule
\parbox[t]{2mm}{\multirow{5}{*}{\rotatebox[origin=c]{90}{\textbf{Q2-Audio}}}}
& 0  & .92 & .63 & .50 & .19 \\
& 8  & .92 & .61 & .49 & .20 \\
& 16 & .90 & .57 & .51 & .19 \\
& 24 & .87 & .49 & .43 & .18 \\
& 31 & .86 & .48 & .29 & .15 \\
\midrule
\parbox[t]{2mm}{\multirow{5}{*}{\rotatebox[origin=c]{90}{\textbf{Ultravox}}}}
& 0  & .86 & .37 & .03 & .07 \\
& 8  & .86 & .35 & .00 & .12 \\
& 16 & .85 & .36 & .07 & .16 \\
& 24 & .83 & .34 & .11 & .15 \\
& 31 & .83 & .31 & .20 & .12 \\
\midrule
\parbox[t]{2mm}{\multirow{5}{*}{\rotatebox[origin=c]{90}{\textbf{Whisper}}}}
& 0  & .99 & .71 & .26 & .06 \\
& 8  & .98 & .78 & .19 & .09 \\
& 16 & .98 & .76 & .16 & .11 \\
& 24 & .97 & .73 & .19 & .12 \\
& 31 & .96 & .67 & .26 & .14 \\
\bottomrule
\end{tabular}
\end{table}

\subsection{Logit Lens: Visualizing Text Emergence}

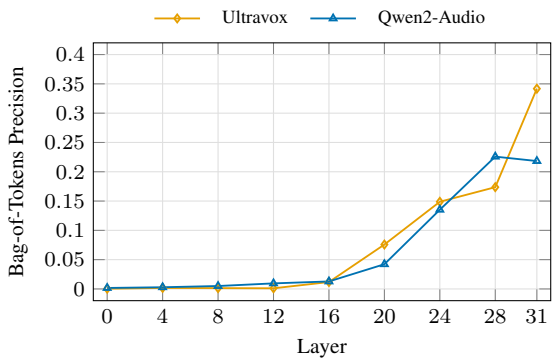
\begin{figure}[!htb]
    \centering
\begin{tikzpicture}[font=\footnotesize]
\begin{axis}[
    width=0.95\columnwidth,
    height=5cm,
    xlabel={Layer},
    ylabel={Bag-of-Tokens Precision},
    xmin=-1, xmax=32,
    xtick={0,4,8,12,16,20,24,28,31},
    ymin=-0.02, ymax=0.42,
    ytick={0,0.05,0.10,0.15,0.20,0.25,0.30,0.35,0.40},
    yticklabel style={/pgf/number format/fixed},
    grid=major,
    grid style={gray!25},
    every axis plot/.append style={line width=0.7pt, mark size=1.5pt},
    legend style={
        at={(0.5,1.02)},
        anchor=south,
        font=\scriptsize,
        legend columns=2,
        column sep=6pt,
        draw=none,
        fill=none,
    },
    clip=false,
]

\addplot[cborange, mark=diamond] coordinates {
  (0,0.0004)(4,0.0017)(8,0.0014)(12,0.0011)(16,0.0118)(20,0.0758)(24,0.1488)(28,0.1737)(31,0.3418)
};
\addlegendentry{Ultravox}

\addplot[cbblue, mark=triangle] coordinates {
  (0,0.0017)(4,0.0029)(8,0.0050)(12,0.0095)(16,0.0128)(20,0.0422)(24,0.1352)(28,0.2258)(31,0.2183)
};
\addlegendentry{Qwen2-Audio}

\end{axis}
\end{tikzpicture}
    \caption{Logit lens: mean bag-of-tokens precision by layer (with RMSNorm applied before projection).
    Both models show text emergence from L20 onward.
    Ultravox (diamonds) surges to $0.34$ at L31; Qwen2-Audio (triangles) peaks at $0.23$ (L28).
    The degree of text emergence mirrors the behavioral agreement spectrum.}
    \label{fig:logit_lens}
\end{figure}

Logit lens~\cite{nostalgebraist2020logitlens,belrose2023tuned} reveals text emergence through the model's own unembedding matrix, requiring no training, unlike probing which requires a trained classifier.
Figure~\ref{fig:logit_lens} and Table~\ref{tab:logit_lens} show bag-of-tokens precision (fraction of greedily decoded tokens appearing in the reference transcript) by layer.

While both Ultravox and Qwen2-Audio do represent text to some extent, there’s a very significant gap between them. 
Ultravox reaches $0.34$ bag precision at layer 31 with recognizable paraphrases (e.g., ``the House of Commons in July'' for a reference containing ``House of Commons on April 27th 1992'').
Qwen2-Audio peaks at $0.23$ (layer 28) with fragmented, multilingual output bearing artifacts from Qwen2's vocabulary.
This gap in text emergence mirrors the behavioral spectrum: the model with stronger text emergence (Ultravox, $\kappa{=}0.75$--$0.93$) shows consistently higher cascade agreement than the model with weaker emergence (Qwen2-Audio, $\kappa{=}0.54$--$0.85$).

\begin{table}[t]
\centering
\caption{Logit lens bag-of-tokens precision by layer.
Fraction of greedy-decoded tokens appearing in the reference transcript ($n{=}1{,}000$).}
\label{tab:logit_lens}
\footnotesize
\setlength{\tabcolsep}{4pt}
\begin{tabular}{rcc}
\toprule
\textbf{Layer} & \textbf{Ultravox} & \textbf{Q2-Audio} \\
\midrule
0  & .000 & .002 \\
4  & .002 & .003 \\
8  & .001 & .005 \\
12 & .001 & .010 \\
16 & .012 & .013 \\
20 & .076 & .042 \\
24 & .149 & .135 \\
28 & .174 & .226 \\
31 & .342 & .218 \\
\bottomrule
\end{tabular}
\end{table}

\subsection{Is the Emergent Text Behaviorally Sufficient?}

The logit lens shows text emerging at audio positions in Ultravox by layer~31.
However, does this text just happen to be a side effect of computation, or is it also the source of information that makes predictions? To verify this on Ultravox, we decode text observed at layer 31 with the logit lens and give it to a second, identical copy of Llama-3.1-8B (the backbone used in Ultravox) with the same prompt. We call this the \emph{implicit cascade}: the text that emerges from the model itself is used as a ``transcript'' to another LLM. If the output of the implicit cascade is equivalent to the behavior of Ultravox, then the text at audio positions is causal with respect to prediction.

Sufficiency of the emergent text is task-dependent, as seen in Table~\ref{tab:implicit}.
On AG News, a topic classification task, the implicit cascade achieves $\kappa_\mathrm{impl} = 0.943$ with Ultravox; \emph{higher} than the agreement between the explicit Whisper cascade and Ultravox ($\kappa_\mathrm{casc} = 0.933$), with only 34\% bag precision. The topic classification task has a low information threshold, a few keywords (``terrorism,'' ``football,'' ``stocks'') are sufficient to classify the topic.
The higher implicit agreement suggests that Ultravox's internal text mirrors its backbone-specific reasoning patterns rather than encoding a generic transcript.
Note that if layer~31 simply encoded the model's final answer, $\kappa$ would be trivially ${\sim}1.0$ on all tasks; instead, the standalone Llama must reason anew over the decoded text.
On SST-2 ($\kappa = 0.14$) and CSQA ($\kappa = 0.25$), the decoded text is insufficient. SST-2 sentiment depends on negation and intensifiers, which are easily lost in noisy decoding; CSQA commonsense reasoning relies on precise relational structure. 
Finally, MELD ($\kappa = 0.15$) serves as a negative control: emotion requires paralinguistic cues absent from any text representation.

\begin{table}[t]
\centering
\caption{Behavioral sufficiency of emergent text at layer 31.
Agreement between implicit cascade and Ultravox ($\kappa_\mathrm{impl}$) and between Whisper cascade and Ultravox ($\kappa_\mathrm{casc}$).}
\label{tab:implicit}
\footnotesize
\begin{tabular}{lrrrr}
\toprule
\textbf{Task} & $\kappa_\mathrm{impl}$ & $\kappa_\mathrm{casc}$ & \textbf{Acc.\ impl} & \textbf{Acc.\ UV} \\
\midrule
AG News & \textbf{0.943} & 0.933 & 88.1 & 86.7 \\
SST-2   & 0.141 & 0.737 & 51.3 & 84.2 \\
CSQA    & 0.253 & 0.800 & 33.1 & 66.9 \\
MELD    & 0.154 & 0.556 & 22.4 & 42.1 \\
\bottomrule
\end{tabular}
\end{table}

\subsection{LEACE}

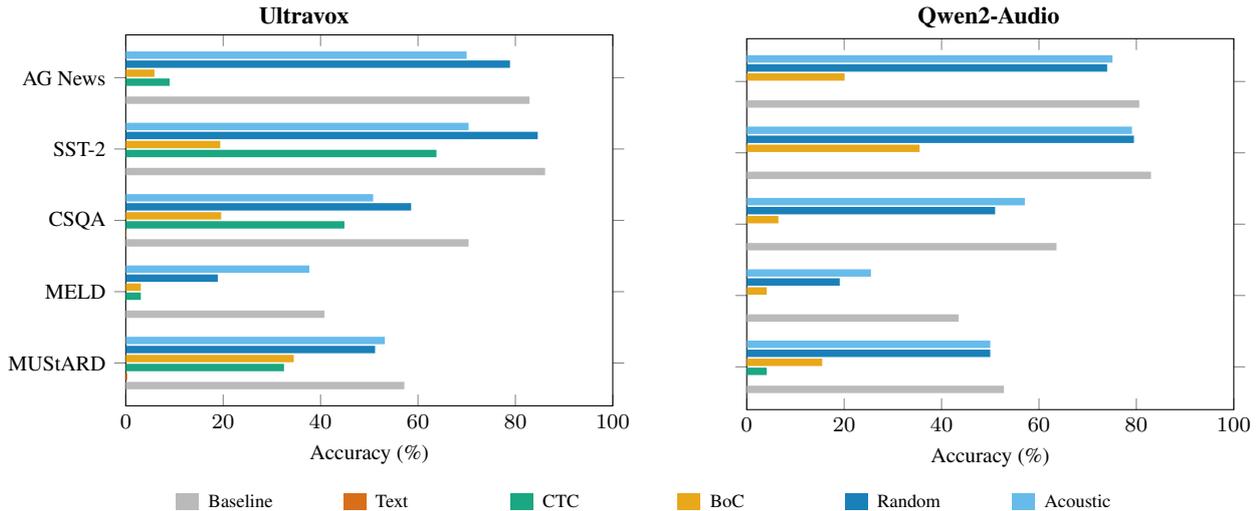
\begin{figure*}[!tb]
    \centering
\begin{minipage}[t]{0.47\textwidth}
\centering
\textbf{Ultravox}\par\smallskip
\begin{tikzpicture}[font=\footnotesize]
\begin{axis}[
    xbar=0.6pt,
    bar width=2.8pt,
    width=\textwidth,
    height=6.5cm,
    xlabel={Accuracy (\%)},
    xmin=0, xmax=100,
    xtick={0,20,40,60,80,100},
    ytick=data,
    symbolic y coords={MUStARD, MELD, CSQA, SST-2, AG News},
    enlarge y limits=0.15,
    clip=false,
    every axis plot/.append style={fill opacity=0.90, draw=none},
]

\addplot[fill=black!30]
    coordinates {(82.9,AG News) (86.1,SST-2) (70.4,CSQA) (40.8,MELD) (57.2,MUStARD)};
\addplot[fill=wongvermillion]
    coordinates {(0.0,AG News) (0.0,SST-2) (0.1,CSQA) (0.0,MELD) (0.3,MUStARD)};
\addplot[fill=wonggreen]
    coordinates {(9.0,AG News) (63.8,SST-2) (44.9,CSQA) (3.1,MELD) (32.5,MUStARD)};
\addplot[fill=wongorange]
    coordinates {(5.9,AG News) (19.4,SST-2) (19.6,CSQA) (3.1,MELD) (34.5,MUStARD)};
\addplot[fill=wongblue]
    coordinates {(78.9,AG News) (84.6,SST-2) (58.6,CSQA) (18.9,MELD) (51.2,MUStARD)};
\addplot[fill=wongsky]
    coordinates {(70.0,AG News) (70.4,SST-2) (50.8,CSQA) (37.7,MELD) (53.2,MUStARD)};

\end{axis}
\end{tikzpicture}
\end{minipage}
\hfill
\begin{minipage}[t]{0.47\textwidth}
\centering
\textbf{Qwen2-Audio}\par\smallskip
\begin{tikzpicture}[font=\footnotesize]
\begin{axis}[
    xbar=0.6pt,
    bar width=2.8pt,
    width=\textwidth,
    height=6.5cm,
    xlabel={Accuracy (\%)},
    xmin=0, xmax=100,
    xtick={0,20,40,60,80,100},
    ytick=data,
    symbolic y coords={MUStARD, MELD, CSQA, SST-2, AG News},
    yticklabels={,,,,},
    enlarge y limits=0.15,
    clip=false,
    every axis plot/.append style={fill opacity=0.90, draw=none},
]

\addplot[fill=black!30]
    coordinates {(80.6,AG News) (83.0,SST-2) (63.6,CSQA) (43.5,MELD) (52.8,MUStARD)};
\addplot[fill=wongvermillion]
    coordinates {(0.0,AG News) (0.0,SST-2) (0.0,CSQA) (0.0,MELD) (0.0,MUStARD)};
\addplot[fill=wonggreen]
    coordinates {(0.0,AG News) (0.0,SST-2) (0.0,CSQA) (0.0,MELD) (4.1,MUStARD)};
\addplot[fill=wongorange]
    coordinates {(20.1,AG News) (35.5,SST-2) (6.5,CSQA) (4.1,MELD) (15.5,MUStARD)};
\addplot[fill=wongblue]
    coordinates {(74.0,AG News) (79.5,SST-2) (51.0,CSQA) (19.1,MELD) (50.0,MUStARD)};
\addplot[fill=wongsky]
    coordinates {(75.1,AG News) (79.1,SST-2) (57.1,CSQA) (25.5,MELD) (50.0,MUStARD)};

\end{axis}
\end{tikzpicture}
\end{minipage}

\vspace{2mm}
\centering
\begin{tikzpicture}[font=\scriptsize]
\newcommand{\lw}{3mm}
\newcommand{\lh}{2.2mm}
\newcommand{\lgap}{1.2mm}
\newcommand{\lsep}{4.5mm}
\foreach \col/\lab/\xoff in {%
    black!30/Baseline/0,%
    wongvermillion/Text/1,%
    wonggreen/CTC/2,%
    wongorange/BoC/3,%
    wongblue/Random/4,%
    wongsky/Acoustic/5%
}{
    \pgfmathsetmacro{\xpos}{\xoff * 22}
    \fill[\col, fill opacity=0.90] (\xpos mm, 0) rectangle +(\lw, \lh);
    \node[right, inner sep=0pt] at (\xpos mm + \lw + \lgap, \lh/2) {\lab};
}
\end{tikzpicture}
    \caption{LEACE results for Ultravox (left) and Qwen2-Audio (right).
    Text erasure collapses both models to near-zero.
    CTC and BoC erasure show an architecture-dependent pattern: CTC is essentially zero for Qwen2-Audio (matching text erasure) but less so for Ultravox, while BoC is more destructive for Ultravox.
    Controls (random, acoustic) confirm specificity.}
    \label{fig:leace}
\end{figure*}

The above evaluations indicate that text representations are \emph{present} (logit lens) and can be \emph{sufficient} (implicit cascade test). LEACE tests causal necessity: does surgically removing text information break task performance? We evaluate LEACE on Ultravox and Qwen2-Audio (Table~\ref{tab:leace}, Figure~\ref{fig:leace}).

Text erasure collapses both models to near-zero accuracy on every task (Ultravox 0.0--0.3\%; Qwen2-Audio 0.0\%), regardless of architecture.
Random erasure of matched dimensionality has negligible effect on text-sufficient tasks (Ultravox AG\,News $-4.0$\%, SST-2 $-1.5$\%), confirming the collapse is text-specific. Text-predictive directions are causally necessary for task performance.

CTC and BoC probes both target text but capture different structure: CTC relies on frame-aligned character sequences, BoC on unordered character frequencies. The LEACE results reveal an architecture-dependent pattern. For Qwen2-Audio, CTC erasure is devastating (0.0\% on four of five tasks) while BoC leaves residual function (AG News 20.1\%, SST-2 35.5\%). For Ultravox, the pattern reverses: BoC is more destructive than CTC (AG News 5.9\% vs. 9.0\%; SST-2 19.4\% vs. 63.8\%). This is consistent with the probing results from \S\ref{sec:layer-wise-probe}: Qwen2-Audio's cross-attention encoder delivers frame-aligned text representations from L0 (CTC TD$=$0.50), making CTC erasure effective; Ultravox's connector produces distributed representations with minimal frame-level text (CTC TD$=$0.03 at L0), so the CTC probe cannot capture, and therefore cannot erase, what it cannot read.

Finally, acoustic erasure (2D) causes unexpected accuracy drops even on text-sufficient tasks.
A diagnostic reveals that in Qwen2-Audio, the acoustic eraser also damages text representations (TD drops by up to $0.25$ at layer~8), confirming acoustic and text subspaces are entangled in cross-attention.
In Ultravox, acoustic erasure does not affect text decodability (TD change $<0.03$), suggesting it disrupts other processing rather than text itself.
MELD performance drops under both random and acoustic erasure in both models, which suggests that emotion representations are fragile to any perturbation; MELD-specific LEACE effects should therefore be interpreted cautiously.

\begin{table}[t]
\centering
\caption{LEACE accuracy (\%) on Ultravox and Qwen2-Audio.
Multi-layer erasure at all 9 probed layers simultaneously;
$n{=}1{,}000$ per task ($n{=}690$ for MUStARD).
$d$: dimensionality of erased subspace.}
\label{tab:leace}
\footnotesize
\setlength{\tabcolsep}{2pt}
\begin{tabular}{llrrrrr}
\toprule
\textbf{Condition} & $d$ & \textbf{AG} & \textbf{SST} & \textbf{CSQA} & \textbf{MELD} & \textbf{MUS} \\
\midrule
\multicolumn{7}{l}{\textit{Ultravox}} \\
Baseline   & ---  & 82.9 & 86.1 & 70.4 & 40.8 & 57.2 \\
Text       & 159  &  0.0 &  0.0 &  0.1 &  0.0 &  0.3 \\
CTC        & 49   &  9.0 & 63.8 & 44.9 &  3.1 & 32.5 \\
BoC        & 48   &  5.9 & 19.4 & 19.6 &  3.1 & 34.5 \\
Acoustic   & 2    & 70.0 & 70.4 & 50.8 & 37.7 & 53.2 \\
Random     & 159  & 78.9 & 84.6 & 58.6 & 18.9 & 51.2 \\
\midrule
\multicolumn{7}{l}{\textit{Qwen2-Audio}} \\
Baseline   & ---  & 80.6 & 83.0 & 63.6 & 43.5 & 52.8 \\
Text       & 159  &  0.0 &  0.0 &  0.0 &  0.0 &  0.0 \\
CTC        & 49   &  0.0 &  0.0 &  0.0 &  0.0 &  4.1 \\
BoC        & 48   & 20.1 & 35.5 &  6.5 &  4.1 & 15.5 \\
Acoustic   & 2    & 75.1 & 79.1 & 57.1 & 25.5 & 50.0 \\
Random     & 159  & 74.0 & 79.5 & 51.0 & 19.1 & 50.0 \\
\bottomrule
\end{tabular}
\end{table}

\subsection{Interpretation}

Speech LLMs build up an internal text representation that determines downstream performance. Probing shows acoustic information is retained but progressively degrades as internal text representations emerge; and while text decodability profile is different across Ultravox, Qwen2-Audio and the Whisper control, all three converge to similar final-layer CTC decodability (0.20–0.29), suggesting that the endpoint is constrained even if the path is not. Logit lens, through a different mechanism, reveals a similar emergence of text, and the gap between Ultravox and Qwen2-Audio mirrors the behavioral ranking of cascade equivalence (Ultravox $>$ Qwen2-Audio). Finally, LEACE demonstrates that this behavior is not a coincidence; we can actively erase these text representations and show a complete collapse in subsequent accuracy. This result is underscored by the fact that matched random erasure does not result in a similar collapse in accuracy. These results indicate that current speech LLMs implement an implicit transcription stage whose structural similarity to ASR output predicts behavioral cascade equivalence.

\section{Discussion}

\textit{When to use cascades vs.\ end-to-end models.}
In a text-sufficient task environment under clean setting, cascades offer performance similar to the state-of-the-art speech LLMs and offer lower latency, cost, and engineering complexity, with separately upgradeable modules. 

In noisy environments, our Whisper-based cascades are more resilient, losing only 0.5--4.2\% at 0\,dB SNR vs.\ 3.9--12.7\% for the E2E models tested. We note that while this advantage has been demonstrated with Whisper's noise-hardened front-end, it should generalize to most modern ASR systems given the widespread adoption of Whisper-adjacent modeling approaches.
The case for E2E architectures rests on tasks that need more than a transcription (e.g., emotion, sarcasm, speaker intent), but even there, probing reveals a gap between retaining and \emph{exploiting} acoustic features.

\textit{Breaking cascade equivalence.} Our results identify the precise bottleneck: the models have the acoustic information but do not use it.
Pitch and energy are retained through the final layers ($R^2 > 0.29$), yet erasing these directions barely affects Ultravox on natural-speech tasks (MELD $-3.1$\%, MUStARD $-4.0$\%). The model does not use information that it has access to. Qwen2-Audio shows a larger effect from acoustic erasure (MELD $-18.0$\%), suggesting its cross-attention architecture partially exploits acoustic information, consistent with its lower cascade equivalence and higher behavioral divergence from its cascade. In both models the information is available but largely unused. This observation leads directly to a solution: we can make the models more responsive to the acoustic information by directly targeting this information in training. 

Potential approaches include: (a) \emph{paralinguistic auxiliary losses} by adding a contrastive loss between audio and text, penalizing the model if it is able to recover audio features from text features. Architectures like WavLLM~\cite{hu2024wavllm}, which explicitly separate semantic and paralinguistic pathways, provide a structural foundation. (b) \emph{minimal-pair prosodic training}: training on audio pairs that differ only in prosody (e.g., sincere vs.\ sarcastic readings of the same sentence) might encourage the model to rely less on the text and more on paralinguistic information. Building on contrastive objectives from self-supervised speech learning~\cite{baevski2020wav2vec}, a loss that pulls apart representations of prosodically distinct utterances with identical transcripts would directly target the textual bottleneck. (c) \emph{architectural modality separation} by employing cross-attention or dual-stream models with anatomically separated acoustic pathways for text and prosody to enable true multimodal inference. Finally, while not as targeted as the previous suggestions, multi-task training across both text sufficient and text insufficient tasks, provides a low cost approach using existing public corpora.

\textit{Implications for benchmarking.}
We demonstrate that a clear confound emerges if tests do not incorporate backbone matching to isolate architectural effects, per-example behavioral analysis beyond aggregate accuracy, and conditions that actually stress the audio pathway. Aggregate benchmarks simply do not capture the underlying capabilities and performance of speech LLMs.

A benchmark suite designed to explore these facets of speech LLMs should include: (1) paralinguistic tasks where the label depends on \emph{how} something is said, not just \emph{what} is said; (2) noisy and reverberant conditions that stress the audio front-end; and (3) matched-backbone cascade baselines to control for LLM reasoning differences.
In the absence of these considerations, it is challenging to assert that architectural benefits in E2E LLMs are genuine.

\textit{Limitations.}
TTS-generated audio has less natural prosody than spontaneous speech, which may inflate acoustic erasure effects: the probe may capture synthesis artifacts rather than genuine paralinguistic information, explaining why acoustic erasure drops are larger on TTS tasks (Ultravox SST-2 $-15.7$\%) than on natural-speech tasks (Ultravox MELD $-3.1$\%). Both linear probing and LEACE are limited to the linear subspace, and would miss any information encoded in a nonlinear fashion. The mechanistic analysis only applied to two of the four E2E models; Gemini is only accessible as an API, and Phi-4-MM was not probed due to its radically different architecture (Mixture-of-LoRAs with a separate conformer encoder), so the mechanistic findings may not extend to all speech LLM architectures.

\section{Conclusion}

We presented the first systematic evaluation of whether speech LLMs are computationally equivalent to ASR$\rightarrow$LLM cascades, using matched-backbone behavioral testing alongside mechanistic evidence on two architectures: Ultravox and Qwen2-Audio.
Our results indicate that speech LLMs behave as implicit ASR$\rightarrow$LLM cascades rather than acoustically grounded end-to-end systems: Ultravox is near-indistinguishable from its matched cascade. Using logit lens and LEACE we confirmed that both architectures rely on text representations to make decisions. However, architecture is still relevant: Qwen2-Audio behaves quite differently; and in the presence of background noise, the cascades outperform end-to-end models.

First, end-to-end models must be evaluated against matched backbones to prevent incorrect conclusions on value-add. Second, for text-sufficient tasks, cascades offer performance and modularity benefits over end-to-end speech LLMs. Finally, end-to-end speech LLMs' promise of better fidelity to both the content and manner of spoken speech is not a mirage, current models retain paralinguistic features but fail to use them, suggesting that training objectives, not architectures, are the bottleneck.

Insofar as our training goals do not prioritize audio-specific cues, speech LLMs will remain cascades in disguise.
Teams developing these models should consider their target applications -- if the application is text-insufficient they must actively incorporate paralinguistic training; for text-sufficient applications the modularity, robustness, and cost benefits of an explicit cascade may outweigh the elegance of an end-to-end speech LLM.

\noindent\textbf{Generative AI Use Disclosure.}
\ifcameraready
  The author used
\else
  The authors used
\fi
generative AI tools (Claude / Claude Code) for manuscript critique, language
polishing and clarity/structure suggestions, and for code debugging/implementation
support. All research design, experiments, analysis, interpretation, manuscript
drafting and revisions are
\ifcameraready
  the author's own.
\else
  the authors' own.
\fi

\bibliographystyle{IEEEtran}
\bibliography{references}

\end{document}